\pdfoutput=1

\documentclass[11pt]{article}

\usepackage[final]{acl}

\usepackage{times}
\usepackage{latexsym}
\usepackage{todonotes}
\usepackage{multirow}
\usepackage{booktabs}
\usepackage{comment}
\usepackage{CJKutf8}
\usepackage[T1]{fontenc}

\usepackage[utf8]{inputenc}

\usepackage{microtype}

\usepackage{inconsolata}

\usepackage{graphicx}
\newcommand{\equalcontrib}{\textsuperscript{*}}
\title{What do Large Language Models Need for \\ Machine Translation Evaluation?} 

\author{Shenbin Qian\textsuperscript{1}, Archchana Sindhujan\textsuperscript{2}\equalcontrib, Minnie Kabra\textsuperscript{3}\equalcontrib \\ \textbf{Diptesh Kanojia\textsuperscript{2}, Constantin Orăsan\textsuperscript{1}, Tharindu Ranasinghe\textsuperscript{4} and Frédéric Blain\textsuperscript{5}} \\
\textsuperscript{1}Centre for Translation Studies, University of Surrey, United Kingdom, \\
\textsuperscript{2}Institute for People-Centred AI, University of Surrey, United Kingdom, \\
\textsuperscript{3}Independent Researcher, India,\\ 
\textsuperscript{4}Lancaster University, United Kingdom,  
\textsuperscript{5}Tilburg University, The Netherlands \\
\{s.qian, a.sindhujan, d.kanojia, c.orasan\}@surrey.ac.uk, minniekabra@gmail.com, \\ t.ranasinghe@lancaster.ac.uk, f.l.g.blain@tilburguniversity.edu
}

\begin{document}
\maketitle

\begingroup
\renewcommand{\thefootnote}{\fnsymbol{footnote}}
\footnotetext[1]{Both authors contributed equally to this work.}
\endgroup

\begin{abstract}
Leveraging large language models (LLMs) for various natural language processing tasks has led to superlative claims about their performance. For the evaluation of machine translation (MT), existing research shows that LLMs are able to achieve results comparable to fine-tuned multilingual pre-trained language models. In this paper, we explore what translation information, such as the source, reference, translation errors and annotation guidelines, is needed for LLMs to evaluate MT quality. In addition, we 
investigate prompting techniques such as zero-shot, Chain of Thought (CoT) and few-shot prompting for eight language pairs covering high-, medium- and low-resource languages, leveraging varying LLM variants. Our findings indicate the importance of reference translations for 
an LLM-based evaluation. While larger models do not necessarily fare better, they tend to benefit more from CoT prompting, than smaller models. We also observe that LLMs do not always provide a numerical score when generating evaluations, which poses a question on their reliability for the task. Our work presents a comprehensive analysis for resource-constrained and training-less LLM-based evaluation of machine translation. We release the accrued prompt templates, code and data publicly for reproducibility\footnote{\url{https://github.com/surrey-nlp/LLM4MT\_eval}}.  

\end{abstract}

\section{Introduction}

Recent surge in the use of large language models (LLMs) for natural language processing (NLP) tasks like question answering~\citep{kocon2023,tan2023} has taken strides, and significantly improved their applications to other downstream tasks such as machine translation (MT), text summarization, information retrieval and~\textit{etc.}, due to advancements in natural language understanding capabilities, contextual awareness, and a versatile knowledge base~\citep{kocmi-federmann-2023-large,zhu2023,zhang2024}. 

For automatic evaluation of MT quality, traditional approaches use metrics such as BLEU~\citep{papineni-etal-2002-bleu}, BLEURT~\citep{sellam-etal-2020-bleurt} or BERTScore~\citep{Zhang2019} to compare MT output with a reference translation. When references are not available, quality estimation (QE) methods such as fine-tuning multilingual pre-trained language models (PTLMs) on human evaluation data like Direct Assessment (DA) scores~\citep{graham-etal-2013-continuous} are often used to predict estimated scores to approximate human evaluation~\citep{Specia2018}. Recent studies leverage prompting techniques and \textit{instruct} LLMs to output a score for translation quality, claim to achieve promising results~\citep{kocmi-federmann-2023-large, kocmi-federmann-2023-gemba}. 

However, there exists no systematic exploration of what translation information LLMs need for quality evaluation, and whether different prompting techniques, such as Chain-of-Thought (CoT)~\citep{Wei2022} or few-shot prompting, can help boost the performance of LLMs. To that end, we conduct this investigation to systematically explore the ability of LLMs in quality evaluation in a training-less scenario. Our contributions can be summarized as: 

\begin{itemize}
    \item We investigate what translation information,~\textit{i.e.}, source, reference, translation errors and annotation guidelines LLMs need to evaluate translation for $8$ language pairs covering high-, medium- and low-resource languages.
    \item We explore different ways of prompting,~\textit{i.e.}, zero-shot, CoT and few-shot prompting for LLMs to evaluate MT quality. Our code, data and prompts are released with the paper.
    \item We compare our prompting methods with fine-tuning of encoder-based multilingual PTLMs and find LLM performance still lags behind.
    \item Our analyses of the results on various prompt templates indicate that
    references are important for accurate translation evaluation with LLMs, and while larger models are not always better, they tend to benefit more from CoT prompting than smaller model variants.
\end{itemize}

The rest of the paper is structured as follows: Section~\ref{lit} discusses relevant work in quality evaluation, while Section~\ref{data} introduces the dataset we utilize for this work. Section~\ref{method} describes the prompting methods and the baselines with the experimental setup. Results and discussion are presented in Section~\ref{results}. Section~\ref{conclusion} concludes our study and outlines future directions. 

\section{Related Work} \label{lit}

Traditional automatic MT quality evaluation metrics such as BLEU, BLEURT and BERTScore compare the MT output to one or several references, whilst metrics like Translation Error Rate (TER)~\citep{snover-etal-2006-study} are based on the number of edits required for MT output to become reference, and neither takes semantic variations into account. 


Training supervised machine learning systems on human-annotated data based on metrics such as DA or Multi-dimensional Quality Metrics (MQM)~\citep{Lommel2014} can help predict translation quality without any references~\cite{deoghare-etal-2023-quality}.~\citet{ranasinghe-etal-2020-transquest, ranasinghe-etal-2021-exploratory} proposed the TransQuest framework to utilize the source text and MT output only and fine-tune XLM-RoBERTa-large~\citep{conneau-etal-2020-unsupervised} to predict a DA score as an estimation of translation quality. COMET~\citep{rei-etal-2020-comet, stewart-etal-2020-comet, rei-etal-2022-cometkiwi} was proposed initially to incorporate references along with the source and MT output to train multilingual PTLMs for quality evaluation, but later it also supported reference-less evaluation.~\citet{wan-etal-2022-unite} proposed a unified translation evaluation framework that could include source or reference or both as input for quality evaluation. Various approaches achieved promising results in the QE shared task of the Conference on Machine Translation (WMT)~\citep{specia-etal-2020-findings-wmt,specia-etal-2021-findings,zerva-etal-2022-findings,blain-etal-2023-findings}, however, most require supervision and training~\citep{deoghare-etal-2023-multi,kanojia-etal-2021-pushing}. 

The advent of LLMs prompted its application to translation quality evaluation.~\citet{kocmi-federmann-2023-large} proposed a zero-shot prompting technique, called GEMBA for DA score prediction using GPT-4~\citep{OpenAI2023}, claiming LLMs can achieve performance comparable to state-of-the-art models fine-tuned on multilingual data. Based on the GEMBA prompt, \citet{fernandes-etal-2023-devil} proposed to use LLMs for both DA score prediction and error categorization via fine-tuning to achieve more fine-grained evaluation. Previous research focused on whether LLMs can be better translation evaluators than state-of-the-art models. To the best of our knowledge, only \citet{Huang2024} investigated how LLMs leverage the source and reference for quality evaluation. However, they only perform zero-shot prompting for three language pairs. Our work comprehensively examines factors such as translation errors and annotation guidelines across eight language pairs, eight prompt templates, and three different prompting techniques. 

\section{Data} \label{data}

We utilized the DA score prediction data released with WMT22 QE shared task~\cite{zerva-etal-2022-findings}. This dataset includes the source (mainly from news articles), MT output (from different MT engines) and (post-edited) human references for eight language pairs,~\textit{i.e.,} English-German (\textbf{EN-DE}), English-Marathi (\textbf{EN-MR}), English-Chinese (\textbf{EN-ZH}), Estonian-English (\textbf{ET-EN}), Nepali-English (\textbf{NE-EN}), Romanian-English (\textbf{RO-EN}), Russian-English (\textbf{RU-EN}) and Sinhala-English (\textbf{SI-EN}). For each source-MT pair, the dataset contains a DA score ranging from $0$ to $100$, rated by human annotators for quality assessment. 

To include annotated errors in the MT output into our prompts, we obtained word-level QE data from WMT22, where tokens of the MT output have sequence labels with either ``OK'' or ``BAD'' indicating translation quality at word level. This dataset also involves the above $8$ language pairs, which contains the source, MT output and the tags for translation quality. For each MT output, we extracted the tokens that were tagged as ``BAD'' as error words. 

Since source-MT segments from sentence-level QE data might differ from those of word-level, we compared each source-MT pair in the two datasets and used the overlapping as the main resource of our research. It includes source, MT output, reference translations and error words for the $8$ language pairs, covering high-, medium- and low-resource languages. We present different prompt templates in Section~\ref{zero-shot_method} to selectively include source, reference and error words to test what translation information LLMs need for quality evaluation. 

We split the data into training, validation, and test sets in proportions of $80\%$, $10\%$, and $10\%$ respectively. We inferenced with LLMs on the test set to obtain evaluation results. Training and validation sets were used to sample examples for few-shot learning (see Section~\ref{few-shot_method}). The size of the test set for each language pair can be seen in Table~\ref{tab:test_size}. 

\begin{table}[h]
\centering
\resizebox{7.5cm}{!}{%
\begin{tabular}{ccccccccc}
\toprule
LP & EN-DE & EN-MR & EN-ZH & ET-EN & NE-EN & RO-EN & RU-EN & SI-EN \\
N & $891$ & $2598$ & $890$ & $897$ & $761$ & $867$ & $900$ & $343$ \\
\bottomrule
\end{tabular}%
}
\caption{The number of instances (N) in the \textbf{test set} for each language pair (LP)}
\label{tab:test_size}
\end{table}

\section{Methodology} \label{method}

This section presents the baselines, and our zero-shot, CoT and few-shot prompting methods.

\subsection{Baselines} \label{baselines}

We utilize TransQuest and COMET, two widely used reference-less and reference-based\footnote{COMET also supports reference-less evaluation.} QE frameworks as baselines. For TransQuest, we employed the fine-tuned MonoTransQuest models proposed by~\citet{ranasinghe-etal-2020-transquest-wmt2020} on each language pair except EN-MR. For EN-MR, we used the English to any model released with TransQuest. For COMET, we utilized a fine-tuned multilingual model ``Unbabel/wmt22-comet-da''~\citep{rei-etal-2022-comet} for all language pairs as it does not have models for each language pair.

\subsection{Zero-shot Prompting} \label{zero-shot_method}

For LLMs to predict translation quality, our prompt includes 1) instructions to perform the task such as ``Score the following translation'', and 2) translation information such as source or reference. 

Since~\citet{kocmi-federmann-2023-large} have shown that their prompt template can achieve state-of-the-art performance using GPT-4, we mainly followed their template to create our prompt instruction as shown in Figure~\ref{gemba_instruction}. We used it as our base template and changed slightly different translation information to test what is needed for LLMs to evaluate MT quality. We constructed prompt \textbf{Template 1} containing ``source + MT output'' as translation information, \textbf{Template 2} ``MT output + reference'', \textbf{Template 3} ``source + MT output + reference'' (exact GEMBA prompt), \textbf{Template 4} ``source + MT output + error words'', \textbf{Template 5} ``source + MT output + reference + error words''. 

\begin{figure*}[h]
\centering
  \begin{minipage}{12cm}
    Score the following translation from \{source\_lang\} to \{target\_lang\} with respect to the \{source/human\_reference/error\_words\} on a continuous scale from 0 to 100, where score of zero means "no meaning preserved" and score of one hundred means "perfect meaning and grammar". \\
    \textbf{\{translation\_information\}} \\
    Score:
  \end{minipage}
  \caption{Base Prompt Template}
\label{gemba_instruction}
\end{figure*}

We augmented the base prompt with summarized guidelines used during human evaluation, as \textbf{Template 6} to test if this could help LLMs evaluate MT quality. These guidelines instruct evaluators to give a DA score by considering multiple factors including accuracy, contextual understanding, grammar, syntax and overall readability.

\subsection{CoT Prompting} \label{CoT_method}

Apart from the translation information and guidelines added in the prompt, we also tested whether CoT prompting could improve LLMs' performance by utilizing reasoning-based steps for quality evaluation. We devised \textbf{Template 7} which includes two-step prompts to score MT quality, as shown in Figure~\ref{cot_template}. In the first prompt, we give translation information (including source, MT output and reference) to the LLM and ask it to analyze step by step where the machine translation is different from the reference. In the second prompt, we instruct the LLM to score the machine translation based on its previous output, \textit{i.e.,} the analysis of machine translation based on reference. Instruction to output a score in JSON format is given to ensure it produces the score first, like other templates. 

\begin{figure*}[h]
\centering
  \begin{minipage}{12cm}
    Prompt 1: \\
    You are going to evaluate the quality for \{language\_pair\} translation. You need to think step by step. First read the following source, machine translation and reference translation. Analyze where the machine translation is different from the reference translation. \\
    Source: \{source\_sentence\} \\
    Machine translation: \{target\_sentence\} \\
    Reference translation: \{reference\_translation\} \\
    Prompt 2: \\
    A large language model did an evaluation of machine translation quality for the \{source\_language\} sentence, which is given as below: \{output\_from\_Prompt1\} \\
    Based on your analysis, score the machine translation quality on a continuous scale from 0 to 100, where score of zero means "no meaning preserved" and score of one hundred means "perfect meaning and grammar". Provide the score strictly in JSON format. 
  \end{minipage}
  \caption{Prompt Template 7}
\label{cot_template}
\end{figure*}

\subsection{Few-shot Learning} \label{few-shot_method}

In addition to zero-shot and CoT prompting, we also added $5$ examples based on Template 3, to show how human annotators score machine translations from $0-100$. We split the training and validation sets into 5 buckets for each language pair according to the score ranges of $0-20$, $21-40$, $41-60$, $61-80$, $81-100$. We randomly sampled $1$ example from each range. The selected $5$ examples for each language pair were given before the instruction for scoring as a prefix (see Figure~\ref{few-shot_prefix}) of the base prompt in Figure~\ref{gemba_instruction}. We call this prompt \textbf{Template 8}.

\begin{figure*}[h]
\centering
  \begin{minipage}{12cm}
    You are going to evaluate the quality of machine translation given the source, machine translation and reference translation. The followings are examples of scoring translation quality. \\
    \{5\_examples\}
  \end{minipage}
  \caption{Prefix for the base prompt to create Template 8}
\label{few-shot_prefix}
\end{figure*}

\subsection{Model Selection} \label{model_select}

We chose 6 models from a variety of open-source LLMs according to their size, popularity and type such as mixture of expert (MoE)~\citep{Shazeer2017} and dense models, and based on our compute capability. For 7-billion-parameter models, we selected Llama-2-7B from Meta~\citep{Touvron2023}, Gemma-7B from Google~\citep{gemma2024} and OpenChat3.5 which was trained with mixed-quality data using Conditional Reinforcement Learning from Human Feedback~\citep{Wang2023}. For 13-billion-parameter models, we opted for Llama-2-13B and Qwen1.5-14B which was specifically tested on a diverse set of 12 languages and showed impressive multilingual capabilities~\citep{Qwen2024}. We also included the Mixtral-8x7B model~\citep{Jiang2024} as our MoE model, but due to the limit of our compute capability, we used the activation-aware weight quantized (AWQ) version~\citep{Lin2023}. For all $6$ selected models, we used the instruction-tuned version, \textit{i.e.,} the chat model, for zero-shot, CoT and few-shot inference. Additionally, we experimented with TowerLLM~\citep{tower_llm_2024} for EN-ZH via HuggingFace\footnote{\url{https://huggingface.co/Unbabel/TowerInstruct-7B-v0.1}}, but results are not discussed in the paper because the model output is mostly identical to the input for most instances (see Appendix~\ref{app:icl}). 

\subsection{Experimental Setup}

All our experiments were run using 1 $\times$ NVIDIA A100 40G, 1 $\times$ A40, and 2 $\times$ RTX A5000 GPUs, for different LLM variants. We used vLLM~\citep{Kwon2023} to save inference time. Detailed settings for hyperparameters, formatting and evaluation metrics are provided below.

\paragraph{Hyperparameters} We chose the default hyperparameter settings in vLLM for all our experiments, \textit{i.e.,} $0.8$ as temperature\footnote{We also experimented with $0$ temperature, but we did not observe huge differences in terms of score distribution.}, $0.95$ for top$\_p$. The input sequence length was chosen as 1024 for zero-shot and CoT inference and 3000 for few-shot inference. 

\paragraph{Formatting} As chat models were fine-tuned on certain formats to interact with humans, it is suggested to use the specific format that was used to train the model while inferencing. As vLLM does not support formatting natively, we formatted all our prompt templates before they were fed into the models based on the format of each model in Section~\ref{model_select}. 

\paragraph{Evaluation} Since LLMs usually output a score and some explanations about their evaluation, we used regular expression to extract the score from the LLM output. Spearman correlation\footnote{Pearson's $r$ and Kendall's $\tau$ are in Appendix~\ref{app:icl}.}~\citep{Spearman1904} was used to evaluate how the predicted scores are correlated with the (mean of) human annotated scores. However, not all LLMs would output a score for all the instances. Sometimes, LLMs failed to score the input translation. In such cases, we dropped these instances (denoted as D in Table~\ref{tab:all-zeroshot}) during the process of correlation calculation, but they were \textit{noted as a metric for robustness}.

\section{Results and Discussion} \label{results}

This section displays our baseline and prompting results using existing QE frameworks and LLMs. 

\subsection{Baselines}

Table~\ref{tab:results_baselines} shows our baseline results using TransQuest and COMET. Since TransQuest models were fine-tuned on data from each language pair with the exception of EN-MR, the Spearman correlation scores of these reference-less models, are higher than those of reference-based COMET models. Except EN-DE and EN-MR, the correlation scores for most language pairs are relatively high. 

\begin{table}[]
\centering
\resizebox{6cm}{!}{%
\begin{tabular}{ccc}
\toprule
LP & TransQuest & COMET \\ \hline
EN-DE & 0.3811 & 0.3579 \\
EN-MR & 0.2489 & 0.5135 \\
EN-ZH & 0.6360 & 0.5410 \\
ET-EN & 0.8148 & 0.7018 \\
NE-EN & 0.8034 & 0.6393 \\
RO-EN & 0.8739 & 0.7699 \\
RU-EN & 0.8252 & 0.6482 \\
SI-EN & 0.7233 & 0.5874 \\
\bottomrule
\end{tabular}%
}
\caption{Spearman $\rho$ achieved by models using TransQuest and COMET on each language pair (LP).}
\label{tab:results_baselines}
\end{table}

\subsection{Zero-shot Inference}

\begin{table*}[]
\centering
\scriptsize
\resizebox{15cm}{!}{%
\begin{tabular}{lcccccccccccc}
\toprule
\multirow{2}{*}{LP} & \multicolumn{2}{c}{T1} & \multicolumn{2}{c}{T2} & \multicolumn{2}{c}{T3} & \multicolumn{2}{c}{T4} & \multicolumn{2}{c}{T5} & \multicolumn{2}{c}{T6} \\
 & $\rho$ & D & $\rho$ & D & $\rho$ & D & $\rho$ & \textbf{D} & $\rho$ & \textbf{D} & $\rho$ & D \\
\hline
\multicolumn{13}{c}{OpenChat3.5}\\
 \hline
EN-DE & 0.2258 & 1 & 0.2209 & 2 & 0.2849 & 0 & 0.2599 & 0 & 0.2812 & 0 & \textbf{\underline{0.2960}} & 0 \\
EN-MR & 0.2295 & 3 & 0.3110 & 9 & \underline{0.3546} & 0 & 0.3347 & 0 & 0.3565 & 0 & 0.3446 & 0 \\
EN-ZH & 0.2722 & 0 & 0.2603 & 4 & \underline{0.3995} & 0 & 0.3002 & 0 & 0.3333 & 0 & 0.3635 & 0 \\
ET-EN & 0.5402 & 0 & 0.5798 & 2 & \textbf{\underline{0.6980}} & 0 & 0.5879 & 0 & 0.6700 & 0 & 0.6925 & 0 \\
NE-EN & 0.3784 & 9 & 0.4855 & 25 & 0.5937 & 0 & 0.5008 & 0 & 0.5832 & 0 & \textbf{\underline{0.6073}} & 0 \\
RO-EN & 0.4712 & 2 & 0.5669 & 25 & 0.7294 & 0 & 0.6900 & 0 & 0.7096 & 0 & \textbf{\underline{0.7385}} & 0 \\
RU-EN & 0.5714 & 0 & 0.5320 & 13 & \textbf{\underline{0.6066}} & 0 & 0.5494 & 0 & 0.5322 & 0 & 0.5938 & 0 \\
SI-EN & 0.4120 & 4 & 0.4201 & 7 & \textbf{\underline{0.6034}} & 0 & 0.4364 & 0 & 0.5990 & 0 & 0.5963 & 0 \\

\hline
\multicolumn{13}{c}{Llama-2-7B}\\
 \hline
EN-DE & 0.0663 & 5 & 0.0397 & 18 & 0.0876 & 73 & -0.0166 & 3 & 0.0887 & 2 & \underline{0.0957} & 27 \\
EN-MR & 0.0417 & 26 & 0.0024 & 85 & \underline{0.1255} & 377 & 0.0154 & 2 & 0.0861 & 1 & 0.0943 & 140 \\
EN-ZH & \underline{0.0956} & 4 & 0.0553 & 15 & 0.0946 & 86 & 0.0273 & 2 & 0.0607 & 2 & 0.0791 & 47 \\
ET-EN & 0.0439 & 3 & 0.1643 & 3 & \underline{0.3715} & 54 & -0.0431 & 1 & 0.2527 & 0 & 0.3319 & 31 \\
NE-EN & 0.1825 & 47 & 0.1018 & 7 & 0.2207 & 85 & 0.0461 & 1 & 0.2026 & 2 & \underline{0.2629} & 26 \\
RO-EN & 0.3068 & 0 & 0.1322 & 4 & \underline{0.4514} & 50 & -0.0059 & 1 & 0.2444 & 0 & 0.3619 & 20 \\
RU-EN & 0.1718 & 13 & 0.1389 & 44 & \underline{0.4253} & 64 & -0.0081 & 15 & 0.2170 & 12 & 0.2404 & 24 \\
SI-EN & 0.0801 & 7 & -0.0238 & 11 & 0.2212 & 36 & 0.0639 & 2 & 0.2288 & 2 & \underline{0.2530} & 18 \\
 
\hline
\multicolumn{13}{c}{ Gemma-7B }\\
 \hline
EN-DE & 0.1516 & 1 & 0.1241 & 0 & 0.1624 & 0 & 0.1074 & 0 & \underline{0.1856} & 0 & 0.1820 & 0 \\
EN-MR & \underline{0.3070} & 0 & 0.2332 & 0 & 0.1479 & 0 & 0.1529 & 0 & 0.2177 & 0 & 0.1815 & 0 \\
EN-ZH & 0.2046 & 0 & 0.1362 & 0 & 0.1805 & 0 & \underline{0.2734} & 0 & 0.2444 & 0 & 0.2342 & 0 \\
ET-EN & 0.3490 & 0 & 0.4074 & 0 & 0.3772 & 0 & 0.4125 & 0 & \underline{0.5552} & 0 & 0.5169 & 0 \\
NE-EN & 0.3329 & 0 & 0.2732 & 2 & 0.2921 & 0 & 0.3439 & 0 & \underline{0.4098} & 0 & \underline{0.4098} & 0 \\
RO-EN & \underline{0.6238} & 0 & 0.4393 & 0 & 0.4429 & 0 & 0.5858 & 0 & 0.5816 & 0 & 0.5911 & 0 \\
RU-EN & 0.3265 & 2 & 0.3697 & 13 & 0.4399 & 0 & 0.3709 & 0 & 0.4450 & 0 & \underline{0.5012} & 0 \\
SI-EN & 0.2740 & 0 & 0.2610 & 0 & 0.3519 & 0 & 0.2816 & 0 & \underline{0.3980} & 0 & 0.3741 & 0 \\

\hline
 \multicolumn{13}{c}{ Llama-2-13B }\\
 \hline
EN-DE & -0.0062 & 535 & -0.0092 & 83 & 0.0316 & 118 & 0.0716 & 10 & \underline{0.1161} & 8 & 0.1061 & 123 \\
EN-MR & 0.0229 & 201 & -0.0692 & 282 & 0.0685 & 224 & 0.0193 & 2 & \underline{0.1051} & 2 & 0.1044 & 483 \\
EN-ZH & 0.0002 & 104 & 0.0032 & 78 & \underline{0.1412} & 118 & 0.0821 & 5 & 0.0967 & 2 & 0.0974 & 206 \\
ET-EN & 0.2159 & 268 & 0.0973 & 84 & 0.4042 & 54 & 0.2196 & 3 & 0.3755 & 5 & \underline{0.4392} & 123 \\
NE-EN & 0.0890 & 78 & 0.2337 & 42 & \underline{0.3178} & 76 & 0.1175 & 4 & 0.1259 & 3 & 0.2895 & 138 \\
RO-EN & 0.2787 & 417 & 0.2484 & 67 & 0.4616 & 50 & 0.2661 & 0 & 0.3224 & 0 & \underline{0.5102} & 133 \\
RU-EN & 0.3931 & 216 & 0.1298 & 99 & 0.4074 & 64 & 0.3328 & 25 & 0.3076 & 26 & \underline{0.4422} & 105 \\
SI-EN & 0.0152 & 79 & 0.3020 & 28 & 0.2669 & 32 & 0.0498 & 2 & 0.0928 & 3 & \underline{0.3659} & 58 \\ 

\hline
 \multicolumn{13}{c}{ Qwen1.5-14B }\\
 \hline
EN-DE & 0.1363 & 16 & 0.2286 & 27 & 0.2182 & 12 & 0.1579 & 0 & 0.2245 & 0 & \underline{0.2359} & 20 \\
EN-MR & 0.3011 & 16 & \textbf{\underline{0.3647}} & 48 & 0.3131 & 12 & 0.2151 & 0 & 0.2838 & 0 & 0.3033 & 17 \\
EN-ZH & 0.3758 & 68 & 0.2500 & 30 & 0.4131 & 11 & 0.3166 & 0 & 0.3504 & 1 & \textbf{\underline{0.4367}} & 18 \\
ET-EN & 0.4836 & 86 & 0.5240 & 132 & 0.6467 & 26 & 0.4741 & 0 & 0.5483 & 0 & \underline{0.6516} & 34 \\
NE-EN & 0.3485 & 213 & 0.4777 & 268 & \underline{0.5114} & 33 & 0.3349 & 0 & 0.4466 & 2 & 0.4651 & 29 \\
RO-EN & 0.2201 & 124 & 0.5161 & 124 & \underline{0.7200} & 17 & 0.5569 & 0 & 0.5790 & 0 & 0.6992 & 18 \\
RU-EN & 0.5157 & 27 & 0.5196 & 96 & \underline{0.5597} & 12 & 0.4743 & 0 & 0.5397 & 1 & 0.5547 & 18 \\
SI-EN & 0.3828 & 71 & 0.4691 & 94 & \underline{0.5936} & 7 & 0.2769 & 0 & 0.4091 & 0 & 0.5427 & 16 \\
 
 \hline
 \multicolumn{13}{c}{ Mixtral-8x7B-AWQ }\\
 \hline
EN-DE & 0.0870 & 4 & 0.0607 & 4 & \underline{0.2631} & 1 & 0.1572 & 2 & 0.1930 & 0 & 0.2309 & 0 \\
EN-MR & 0.1067 & 19 & 0.0799 & 22 & 0.1825 & 2 & 0.0872 & 8 & \underline{0.2078} & 8 & 0.1936 & 1 \\
EN-ZH & 0.3390 & 1 & 0.1253 & 1 & \underline{0.3720} & 0 & 0.2104 & 0 & 0.2746 & 5 & 0.3434 & 0 \\
ET-EN & 0.3128 & 9 & 0.2081 & 3 & \underline{0.6229} & 1 & 0.3499 & 3 & 0.4338 & 3 & 0.5903 & 0 \\
NE-EN & 0.4019 & 7 & 0.4025 & 8 & \underline{0.4891} & 0 & 0.1212 & 3 & 0.2279 & 1 & 0.4684 & 0 \\
RO-EN & 0.3204 & 13 & 0.2557 & 16 & \underline{0.6526} & 0 & 0.4053 & 2 & 0.4404 & 2 & 0.6164 & 1 \\
RU-EN & 0.4750 & 4 & 0.3742 & 13 & \underline{0.5831} & 0 & 0.4160 & 1 & 0.5022 & 3 & 0.5528 & 0 \\
SI-EN & 0.2652 & 6 & 0.2139 & 9 & \underline{0.4563} & 0 & 0.0966 & 2 & 0.2220 & 1 & 0.4124 & 1 \\
\bottomrule

\end{tabular}%
}
\caption{Spearman $\rho$ correlation scores achieved by zero-shot inference using \textbf{Templates 1-6 (T1-6)} on various open-source LLMs for each language pair (LP). D -> rows dropped as LLM generated output without a score. The underlined scores represent the best result among templates, while the bold represent the best among LLMs.}
\label{tab:all-zeroshot}
\end{table*}

Table~\ref{tab:all-zeroshot} shows our zero-shot results of Templates 1 to 6 for open-source LLMs including OpenChat3.5, Llama-2-7B, Gemma7B, Llama-2-13B, Qwen1.5-14B and Mixtral-8x7B-AWQ. 


\paragraph{Comparing results among LLMs} We observe that OpenChat3.5 achieved the highest Spearman correlation scores for most language pairs, despite having only $7$ billion parameters – roughly half the size of Llama-2-13B and Qwen1.5-14B. It excelled not only in Spearman scores but also in consistently providing valid quality evaluation scores, with very few dropped rows. Among the 6 models, Llama-2 (both the $7$ and $13$ billion variants) performed poorly in generating evaluations with valid scores. Many rows were dropped, and the Spearman scores were low, indicating a weak correlation with the true scores. The MoE model, Mixtral-8x7B-AWQ, did not outperform OpenChat3.5 on most language pairs and prompt templates for our task.

\paragraph{Comparing with the baselines} We find that models fine-tuned for each language pair by TransQuest, performed much better than the zero-shot prompting results for all language pairs. For some language pairs like ET-EN, NE-EN and RO-EN, our best zero-shot prompting results (Template 6 of OpenChat3.5 in Table~\ref{tab:all-zeroshot}) were comparable to the reference-based models fine-tuned on multilingual data using COMET. For some other pairs like SI-EN, our best zero-shot prompting results were even slightly better than the COMET models. 

\begin{figure*}[h]
  \centering
  \includegraphics[width=0.97\textwidth]{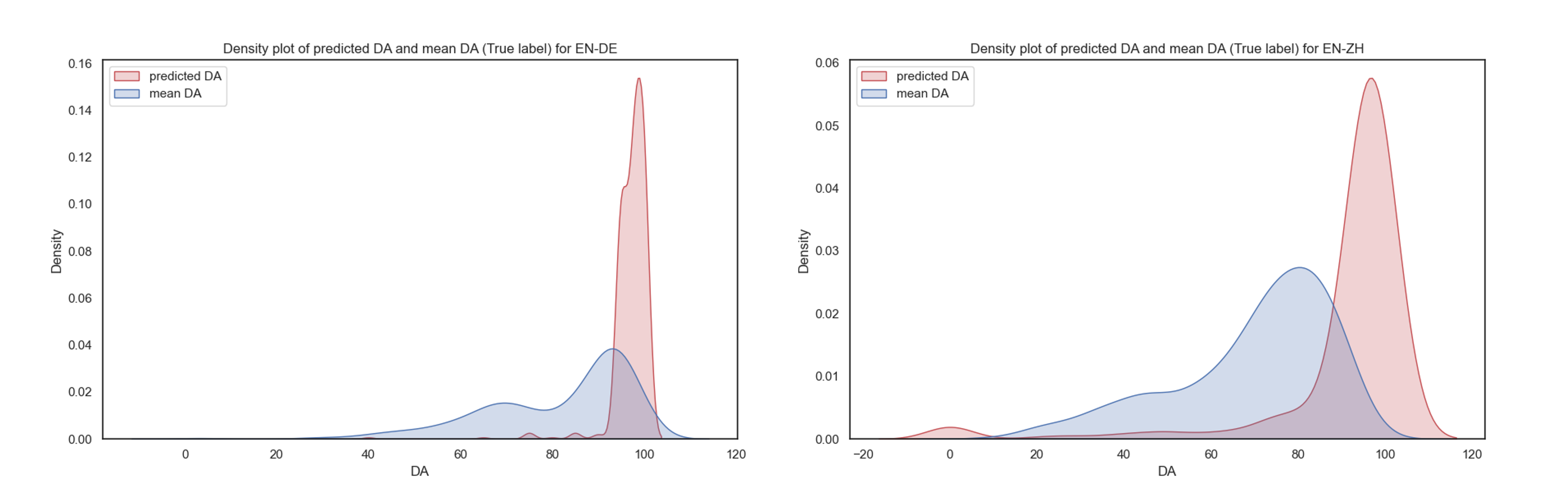}
  \caption{Density plots of the predicted (red) and true (mean) DA scores (blue) for high-resource language pairs~\textit{i.e.}, EN-DE (left) and EN-ZH (right).}
  \label{fig.high}
\end{figure*}

\begin{figure*}[h]
  \centering
  \includegraphics[width=0.97\textwidth]{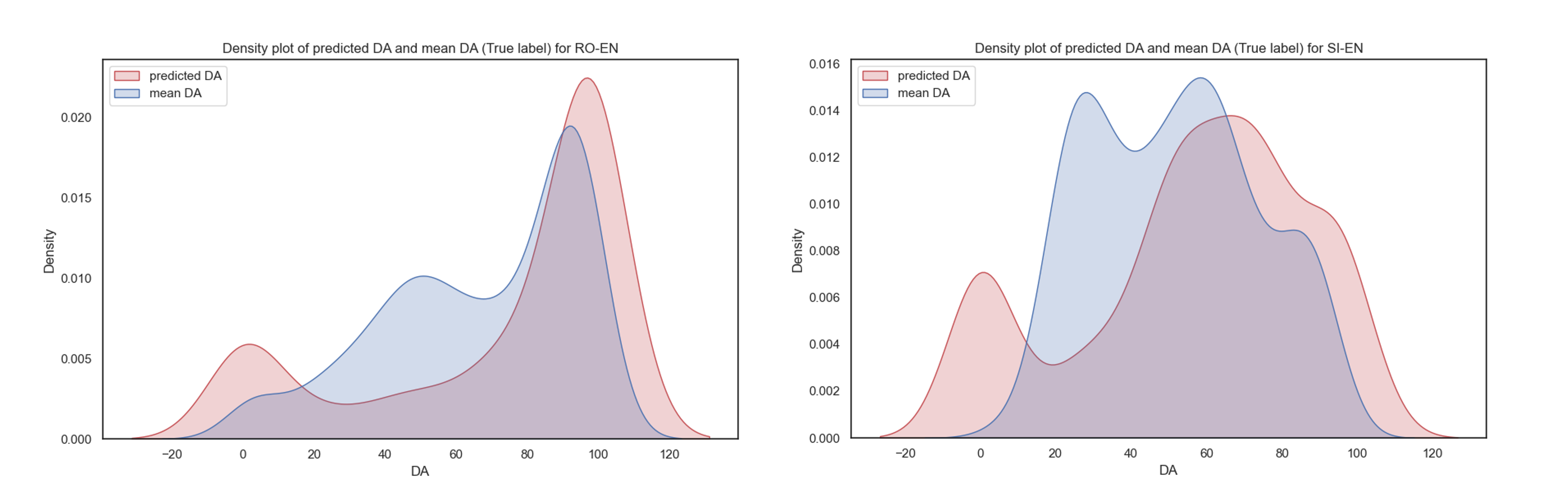}
  \caption{Density plots of the predicted (red) and true (mean) DA scores (blue) for medium- and low-resource language pairs~\textit{i.e.}, RO-EN (left) and SI-EN (right).}
  \label{fig.low}
\end{figure*}

\paragraph{Comparison among Templates} When we fix the model variable as OpenChat3.5, we can compare the performance of different prompt templates. Looking at the OpenChat3.5 results in Table~\ref{tab:all-zeroshot}, we observe that LLM performance is generally better when the source and reference are included in the prompt, as in Templates 3, 5, and 6, compared to prompts without them, such as Templates 1 and 2. This pattern holds true for other LLMs such as the LLama-2 models and Gemma, as shown in the table. Notably, the Spearman scores are obviously higher when the source is incorporated into the prompt, as seen by comparing Templates 2 and 3. This suggests that the source is an essential component for evaluating MT quality using LLMs, contrary to the results in~\citet{Huang2024}, who indicate that the source provides a negative impact. 

Our results (on Templates 4, 5, 6) suggest that including error words and annotation guidelines does not consistently help LLMs evaluate MT quality across different language pairs when compared to using just the plain GEMBA prompt (Template 3). For most language pairs like EN-ZH, ET-EN, RU-EN, and SI-EN, Template 3 had the highest correlation with human judgments. However, removing reference translations (Template 4) clearly lowered correlation scores, highlighting their importance for accurate MT evaluation.

Although incorporating error words does not seem to improve performance, they are surprisingly useful in helping LLMs provide scores in their outputs. As shown in Table~\ref{tab:all-zeroshot},
there are fewer dropped rows when using Templates 4 and 5, which include error words. Outputs from Templates 4 and 5 are the most stable across models, unlike other templates that are more model-dependent.


\paragraph{Results among different language pairs} For high-resource language pairs like EN-DE and EN-ZH, correlation scores tend to be lower than those of medium- and low-resource pairs such as NE-EN, RO-EN, and RU-EN. This pattern holds true across most models, including the fine-tuned ones from TransQuest and COMET.

To further investigate the reasons, we selected EN-DE and EN-ZH as high-resource language pairs, and RO-EN and SI-EN as medium- and low-resource language pairs. We plotted the distributions of the predicted (from OpenChat3.5) \textit{vs} true scores (mean of all annotators) as shown in Figures~\ref{fig.high} and~\ref{fig.low}. For high-resource language pairs, the predictions are skewed towards higher DA scores. Well-trained MT systems, due to abundant resources, tend to produce high-quality translations, leading to higher DA scores. However, LLM-based evaluation systems may amplify these imbalanced distributions and are more likely to predict scores within the high range.

\begin{table*}[h]
\centering
\resizebox{15cm}{!}{%
\begin{tabular}{ccccccccccccc}
\toprule
\multirow{2}{*}{LP} & \multicolumn{2}{c}{OpenChat3.5} & \multicolumn{2}{c}{Llama-2-7B} & \multicolumn{2}{c}{Gemma-7B} & \multicolumn{2}{c}{Llama-2-13B} & \multicolumn{2}{c}{Qwen1.5-14B} & \multicolumn{2}{c}{Mixtral-8x7B-AWQ} \\
 & T7 & T3 & T7 & T3 & T7 & T3 & T7 & T3 & T7 & T3 & T7 & T3 \\ \hline
EN-DE & 0.2433 & \textbf{\underline{0.2849}} & -0.0353 & \underline{0.0876} & 0.0048 & \underline{0.1624} & \underline{0.0345} & 0.0316 & \underline{0.2388} & 0.2182 & 0.2213 & \underline{0.2631} \\
EN-MR & 0.2937 & \textbf{\underline{0.3546}} & -0.0021 & \underline{0.1255} & 0.0859 & \underline{0.1479} & \underline{0.0804} & 0.0685 & \underline{0.3455} & 0.3131 & \underline{0.1906} & 0.1825 \\
EN-ZH & 0.3324 & \textbf{\underline{0.3995}} & 0.0354 & \underline{0.0946} & 0.1609 & \underline{0.1805} & 0.0703 & \underline{0.1412} & 0.3429 & \underline{0.4131} & 0.2479 & \underline{0.3720} \\
ET-EN & 0.6110 & \textbf{\underline{0.6980}} & 0.1459 & \underline{0.3715} & 0.3191 & \underline{0.3772} & 0.2558 & \underline{0.4042} & 0.5845 & \underline{0.6467} & 0.4628 & \underline{0.6229} \\
NE-EN & 0.5160 & \textbf{\underline{0.5937}} & 0.1363 & \underline{0.2207} & \underline{0.3221} & 0.2921 & \underline{0.3315} & 0.3178 & 0.4791 & \underline{0.5114} & 0.4373 & \underline{0.4891} \\
RO-EN & 0.7175 & \textbf{\underline{0.7294}} & 0.1859 & \underline{0.4514} & \underline{0.4550} & 0.4429 & 0.3403 & \underline{0.4616} & 0.7019 & \underline{0.7200} & 0.6360 & \underline{0.6526} \\
RU-EN & 0.5317 & \textbf{\underline{0.6066}} & 0.1618 & \underline{0.4253} & 0.2979 & \underline{0.4399} & 0.2519 & \underline{0.4074} & 0.5203 & \underline{0.5597} & 0.5191 & \underline{0.5831} \\
SI-EN & 0.5124 & \textbf{\underline{0.6034}} & 0.1818 & \underline{0.2212} & 0.2808 & \underline{0.3519} & \underline{0.2854} & 0.2669 & 0.4680 & \underline{0.5936} & \underline{0.4691} & 0.4563 \\
\bottomrule
\end{tabular}%
}
\caption{Spearman $\rho$ correlation scores achieved using \textbf{Template 7 (T7)}, the CoT prompt template, on various LLMs for each language pair (LP). Results of \textbf{Template 3 (T3)} from Table \ref{tab:all-zeroshot} are listed here for reference. The underlined scores represent better result between the templates, while the bold represent the best among LLMs.}
\label{tab:t7t3}
\end{table*}

\begin{table*}[h]
\centering
\resizebox{15cm}{!}{%
\begin{tabular}{ccccccccccccc}
\toprule
\multirow{2}{*}{LP} & \multicolumn{2}{c}{OpenChat3.5} & \multicolumn{2}{c}{Llama-2-7B} & \multicolumn{2}{c}{Gemma-7B} & \multicolumn{2}{c}{Llama-2-13B} & \multicolumn{2}{c}{Qwen1.5-14B} & \multicolumn{2}{c}{Mixtral-8x7B-AWQ} \\
 & T8 & T3 & T8 & T3 & T8 & T3 & T8 & T3 & T8 & T3 & T8 & T3 \\ \hline
EN-DE & 0.1756 & \textbf{\underline{0.2849}} & 0.0327 & \underline{0.0876} & 0.0343 & \underline{0.1624} & \underline{0.0655} & 0.0316 & 0.1804 & \underline{0.2182} & 0.2155 & \underline{0.2631} \\
EN-MR & 0.2543 & \textbf{\underline{0.3546}} & 0.0078 & \underline{0.1255} & \underline{0.1651} & 0.1479 & 0.0159 & \underline{0.068} & 0.2706 & \underline{0.3131} & \underline{0.2410} & 0.1825 \\
EN-ZH & 0.2801 & \underline{0.3995} & 0.0283 & \underline{0.0946} & \underline{0.1831} & 0.1805 & 0.0875 & \underline{0.1412} & 0.2946 & \textbf{\underline{0.4131}} & 0.2970 & \underline{0.3720} \\
ET-EN & 0.5779 & \textbf{\underline{0.6980}} & -0.0026 & \underline{0.3715} & \underline{0.4134} & 0.3772 & 0.2328 & \underline{0.4042} & 0.4320 & \underline{0.6467} & 0.5566 & \underline{0.6229} \\
NE-EN & 0.4621 & \textbf{\underline{0.5937}} & 0.1428 & \underline{0.2207} & \underline{0.3117} & 0.2921 & 0.1907 & \underline{0.3178} & 0.3349 & \underline{0.5114} & \underline{0.5143} & 0.4891 \\
RO-EN & 0.6881 & \textbf{\underline{0.7294}} & 0.0405 & \underline{0.4514} & \underline{0.4693} & 0.4429 & 0.2574 & \underline{0.4616} & 0.4498 & \underline{0.7200} & \underline{0.6712} & 0.6526 \\
RU-EN & 0.5774 & \textbf{\underline{0.6066}} & 0.1680 & \underline{0.4253} & 0.2531 & \underline{0.4399} & 0.1951 & \underline{0.4074} & 0.4798 & \underline{0.5597} & 0.5239 & \underline{0.5831} \\
SI-EN & 0.4277 & \textbf{\underline{0.6034}} & 0.0352 & \underline{0.2212} & 0.3048 & \underline{0.3519} & 0.1368 & \underline{0.2669} & 0.4207 & \underline{0.5936} & \underline{0.4642} & 0.4563 \\
\bottomrule
\end{tabular}%
}
\caption{Spearman $\rho$ correlation scores achieved using \textbf{Template 8 (T8)}, the few-shot prompt template, on various LLMs for each language pair (LP). Results of \textbf{Template 3 (T3)} from Table 3 are listed here for reference. The underlined scores represent better result between the templates, while the bold represent the best among LLMs.}
\label{tab:t8t3}
\end{table*}

In contrast, for medium- and low-resource language pairs, there are fewer resources for training MT systems. As a result, low-quality translations (with low DA scores) are better represented than in high-resource pairs. Quality evaluation systems can better recognize low-quality translations and produce a more balanced score distribution. This imbalance in the score representation could be the reason why predicted DA scores for high-resource languages are less correlated with true scores than for medium- and low-resource pairs.

\subsection{CoT and Few-shot Inference}

Tables~\ref{tab:t7t3} and~\ref{tab:t8t3} show results of CoT (Template 7) and 5-shot inference (Template 8) together with the results of Template 3 for the 6 selected LLMs. Dropped rows for the two templates are presented in Table~\ref{tab:D-t7t8}. Both Templates 7 and 8 were built upon Template 3,~\textit{i.e.}, including the source, MT output and reference. We expect the model performance to be improved when more reasoning steps or evaluation examples were given. However, for $7$ billion parameter variants, CoT prompting resulted in worse performance, as Spearman correlation scores of Template 7 were obviously lower than those of Template 3. For the larger $13$ billion parameter variants, results were mixed for different language pairs. For language pairs such as EN-DE and EN-MR, CoT prompting improved the performance in the prediction of DA scores. This indicates that CoT may work better on larger models than smaller models. While CoT prompting did not consistently improve model performance as measured by the Spearman correlation scores, it shows relatively more consistent output than other prompt templates. Table~\ref{tab:D-t7t8} suggests that fewer rows were dropped when using Template 7, especially for Llama-2 models.

Interestingly, 5-shot inference results are not better than zero-shot results, posing a question on context utilization by LLMs. Performance varies on the LLMs and the specific language pairs. This could relate to the language data available for training these LLMs, as well as the quality of the evaluation examples chosen for different languages pairs.

\begin{table*}[]
\centering
\resizebox{13cm}{!}{%
\begin{tabular}{ccccccccccccc}
\toprule
\multirow{2}{*}{LP} & \multicolumn{2}{c}{OpenChat3.5} & \multicolumn{2}{c}{Llama-2-7B} & \multicolumn{2}{c}{Gemma-7B} & \multicolumn{2}{c}{Llama-2-13B} & \multicolumn{2}{c}{Qwen1.5-14B} & \multicolumn{2}{c}{Mixtral-8x7B-AWQ} \\ 
 & T7 & T8 & T7 & T8 & T7 & T8 & T7 & T8 & T7 & T8 & T7 & T8 \\ \hline
EN-DE & 0 & 0 & 0 & 1 & 0 & 0 & 0 & 0 & 3 & 9 & 0 & 0 \\
EN-MR & 0 & 0 & 0 & 7 & 0 & 0 & 0 & 0 & 1 & 8 & 0 & 2 \\
EN-ZH & 0 & 0 & 0 & 21 & 0 & 1 & 0 & 0 & 1 & 24 & 0 & 0 \\
ET-EN & 0 & 0 & 0 & 0 & 0 & 0 & 0 & 0 & 38 & 22 & 0 & 0 \\
NE-EN & 0 & 0 & 1 & 1 & 0 & 0 & 1 & 0 & 36 & 23 & 0 & 2 \\
RO-EN & 0 & 0 & 0 & 0 & 0 & 1 & 0 & 0 & 27 & 69 & 1 & 0 \\
RU-EN & 0 & 0 & 4 & 7 & 4 & 0 & 3 & 1 & 17 & 29 & 1 & 0 \\
SI-EN & 0 & 0 & 1 & 2 & 0 & 0 & 0 & 2 & 4 & 5 & 1 & 0 \\
\bottomrule
\end{tabular}%
}
\caption{\textbf{D}ropped rows for \textbf{Template 7 (T7)} and \textbf{Template 8 (T8)}, \textit{i.e.}, the CoT and few-shot prompt templates, using various LLMs for each language pair (LP).}
\label{tab:D-t7t8}
\end{table*}

\subsection{Discussion}

Based on our results, \textit{Template 3}, which includes the source, MT output and reference, but excludes error words and detailed guidelines, performed the best in terms of Spearman correlation scores. Prompting with CoT and few-shot learning may yield better results for larger models, but more experiments are needed to confirm this. 

While larger language models often perform better, our results show that a $7$-billion parameter model outperformed other models for most language pairs. Surprisingly, even much smaller COMET models fine-tuned on multilingual data, rather than data for specific language pairs, usually outperformed our LLM prompting results. However, due to the high computational cost, we could not test models with $70$ billion or more parameters.

Different models excel at various language pairs while struggling with others. Even for a single model, performance fluctuates across different language pairs. This variability could stem from whether a language is considered high-resource, but further research is necessary to understand the underlying causes.

Our experiments with prompting LLMs for translation evaluation reveal that these models are often inconsistent in \textit{generating numerical scores}. In most cases, LLMs tend to generate scores accompanied by lengthy and unstructured explanations. While using regular expressions for extraction can be helpful, it is not always reliable. For models like Llama-2, we observed numerous instances where LLMs failed to produce a valid score. Our empirical findings demonstrate that employing CoT prompting or incorporating error words into the prompt can enhance the consistency of the model outputs.


\section{Conclusion and Future Work} \label{conclusion}

In this paper, we explored what translation information is needed for LLMs to evaluate MT quality. We conducted a comprehensive investigation into different prompting techniques such as zero-shot, CoT and few-shot prompting using different translation information for 8 language pairs and 6 LLMs of different sizes and types. Our findings suggest that the source, MT output and reference are essential compared to other information such as translation errors for quality evaluation. Larger models may not necessarily perform better than smaller models, but CoT prompting works better on larger than smaller model variants. We also observe that LLMs do not always provide a numerical score when generating evaluations, which makes their assessments less reliable. For future research, we plan to explore whether fine-tuning LLMs could improve their performance in quality evaluation. We also plan to thoroughly investigate error explainability of LLMs using MQM and other fine-grained error identification techniques. These future studies can inform downstream error correction through automatic post-editing, contributing to a more comprehensive evaluation and correction framework. 

\section*{Limitations and Ethical Considerations} \label{limitations}

Our results were achieved on a limited number of LLMs which are mostly smaller than 14 billion parameters due to the constraints of our computational capabilities. Larger models may perform differently in this translation evaluation task. The examples used in the few-shot scenario were randomly sampled since we do not have the knowledge to prepare good-quality examples for all language pairs. Results might be different if these examples were carefully chosen by native speakers. 

Our experiments in the paper were conducted solely on publicly available datasets as described in Section~\ref{data}, requiring no ethical approval. 

\section*{Acknowledgements} \label{ack}

We would like to thank the European Association for Machine Translation (EAMT) for funding QE data curation of Indic languages used in this paper (UoS/RN0580). 

\bibliography{custom,anthology}

\clearpage 
\onecolumn 
\appendix

\section{ Pearson's $r$ and Kendall's $\tau$ Correlation Scores} \label{app:icl}

\begin{table*}[ht]
\centering
\normalsize 
\resizebox{16cm}{!}{%
\begin{tabular}{cccccccccccccccccccc}
\toprule
\multirow{2}{*}{LP} & \multicolumn{2}{c}{T1} & \multicolumn{2}{c}{T2} & \multicolumn{2}{c}{T3} & \multicolumn{2}{c}{T4} & \multicolumn{2}{c}{T5} & \multicolumn{2}{c}{T6} & \multicolumn{2}{c}{T7} & \multicolumn{2}{c}{T8} \\ \cmidrule(lr){2-3} \cmidrule(lr){4-5} \cmidrule(lr){6-7} \cmidrule(lr){8-9} \cmidrule(lr){10-11} \cmidrule(lr){12-13} \cmidrule(lr){14-15} \cmidrule(lr){16-17}
 & $r$ & $\tau$ & $r$ & $\tau$ & $r$ & $\tau$ & $r$ & $\tau$ & $r$ & $\tau$ & $r$ & $\tau$ & $r$ & $\tau$ & $r$ & $\tau$ \\ \hline
\multicolumn{17}{c}{ \textbf{OpenChat3.5} }\\
 \hline
EN-DE & 0.2048 & 0.1613 & 0.2157 & 0.1556 & 0.2932 & 0.2153 & 0.2305 & 0.1956 & 0.3094 & 0.2135 & 0.3246 & 0.2251 & 0.2180 & 0.1746 & 0.2179 & 0.1279 \\
EN-MR & 0.2551 & 0.2031 & 0.2774 & 0.2023 & 0.5192 & 0.2757 & 0.3351 & 0.2560 & 0.4463 & 0.2711 & 0.4919 & 0.2654 & 0.3669 & 0.2173 & 0.3529 & 0.1791 \\
EN-ZH & 0.2921 & 0.2267 & 0.2655 & 0.1894 & 0.4001 & 0.2905 & 0.3063 & 0.2191 & 0.3702 & 0.2406 & 0.3865 & 0.2617 & 0.3487 & 0.2364 & 0.2971 & 0.1938 \\
ET-EN & 0.5474 & 0.4107 & 0.5966 & 0.4418 & 0.6776 & 0.5249 & 0.5753 & 0.4341 & 0.6617 & 0.5043 & 0.6679 & 0.5213 & 0.5937 & 0.4452 & 0.5610 & 0.4146 \\
NE-EN & 0.3606 & 0.2705 & 0.4968 & 0.3617 & 0.6185 & 0.4364 & 0.5447 & 0.3649 & 0.6114 & 0.4314 & 0.6246 & 0.4478 & 0.5547 & 0.3849 & 0.4809 & 0.3287 \\
RO-EN & 0.4312 & 0.3664 & 0.5307 & 0.3971 & 0.7891 & 0.5642 & 0.7353 & 0.5359 & 0.2487 & 0.5516 & 0.7983 & 0.5716 & 0.7634 & 0.5433 & 0.7491 & 0.5097 \\
RU-EN & 0.5893 & 0.4714 & 0.4760 & 0.3630 & 0.6655 & 0.4761 & 0.4957 & 0.4290 & 0.5926 & 0.4183 & 0.6643 & 0.4630 & 0.6016 & 0.4009 & 0.6413 & 0.4199 \\
SI-EN & 0.4060 & 0.3060 & 0.4351 & 0.3135 & 0.6001 & 0.4492 & 0.4434 & 0.3147 & 0.5957 & 0.4449 & 0.5920 & 0.4395 & 0.5139 & 0.3811 & 0.4165 & 0.2987 \\ 

\hline
\multicolumn{17}{c}{ \textbf{Llama-2-7B}  }\\
 \hline

EN-DE & 0.0436 & 0.0344 & 0.0861 & 0.0642 & 0.0216 & 0.0662 & 0.0494 & -0.0110 & 0.0132 & 0.0693 & -0.0185 & 0.0686 & -0.0225 & -0.0259 & -0.0914 & 0.0235 \\
EN-MR & 0.0763 & 0.0597 & 0.0481 & 0.0361 & -0.0484 & 0.0931 & 0.0271 & 0.0122 & 0.0127 & 0.0647 & 0.0447 & 0.0686 & -0.0096 & -0.0017 & 0.0222 & 0.0056 \\
EN-ZH & 0.0818 & 0.0623 & 0.0478 & 0.0361 & 0.0938 & 0.0712 & -0.0069 & 0.0208 & 0.0121 & 0.0459 & 0.0470 & 0.0579 & 0.0393 & 0.0261 & 0.0368 & 0.0194 \\
ET-EN & 0.1231 & 0.0978 & 0.2589 & 0.1976 & 0.0452 & 0.2805 & 0.0382 & -0.0301 & 0.0593 & 0.1894 & 0.0574 & 0.2418 & 0.1224 & 0.1065 & -0.0529 & -0.0023 \\
NE-EN & 0.2156 & 0.1680 & 0.1729 & 0.1308 & 0.1288 & 0.1670 & 0.0617 & 0.0344 & 0.0164 & 0.1503 & 0.0182 & 0.1939 & 0.1348 & 0.0995 & 0.0311 & 0.0994 \\
RO-EN & 0.2662 & 0.2072 & 0.2364 & 0.1807 & 0.0886 & 0.3404 & -0.0514 & -0.0068 & 0.0571 & 0.1772 & 0.0030 & 0.2622 & 0.0718 & 0.1358 & -0.0357 & 0.0285 \\
RU-EN & 0.2342 & 0.1891 & 0.2084 & 0.1564 & 0.3123 & 0.3234 & 0.0273 & -0.0040 & 0.0270 & 0.1632 & 0.0030 & 0.1759 & 0.1879 & 0.1182 & 0.0531 & 0.1167 \\
SI-EN & 0.1295 & 0.1019 & 0.0382 & 0.0300 & 0.1345 & 0.1638 & 0.0852 & 0.0428 & 0.1594 & 0.1713 & 0.1287 & 0.1867 & 0.1719 & 0.1339 & 0.0725 & 0.0250 \\

\hline
\multicolumn{17}{c}{ \textbf{Gemma-7B}   }\\
 \hline

EN-DE & 0.1217 & 0.0984 & 0.1578 & 0.1239 & 0.1622 & 0.1280 & 0.0994 & 0.0844 & 0.1696 & 0.1436 & 0.1693 & 0.1424 & 0.0193 & 0.0040 & -0.0066 & 0.0244 \\
EN-MR & 0.1745 & 0.1428 & 0.2018 & 0.1599 & 0.2114 & 0.1189 & 0.2576 & 0.1256 & 0.2634 & 0.1747 & 0.2024 & 0.1468 & 0.1379 & 0.0662 & 0.0456 & 0.1182 \\
EN-ZH & 0.2724 & 0.2121 & 0.2037 & 0.1581 & 0.1100 & 0.1406 & 0.2622 & 0.2156 & 0.2091 & 0.1876 & 0.2171 & 0.1826 & 0.1516 & 0.1228 & 0.0372 & 0.1274 \\
ET-EN & 0.3837 & 0.3003 & 0.4749 & 0.3721 & 0.3452 & 0.2954 & 0.3922 & 0.3237 & 0.5009 & 0.4294 & 0.4522 & 0.3979 & 0.3483 & 0.2382 & 0.1742 & 0.2927 \\
NE-EN & 0.3794 & 0.3002 & 0.3242 & 0.2595 & 0.2635 & 0.2244 & 0.3204 & 0.2658 & 0.3677 & 0.3191 & 0.3399 & 0.3111 & 0.3051 & 0.2425 & 0.0448 & 0.2217 \\
RO-EN & 0.5852 & 0.4552 & 0.4672 & 0.3662 & 0.4127 & 0.3473 & 0.6256 & 0.4585 & 0.5523 & 0.4558 & 0.5630 & 0.4600 & 0.5137 & 0.3468 & 0.1365 & 0.3430 \\
RU-EN & 0.4205 & 0.3294 & 0.4479 & 0.3448 & 0.3331 & 0.3452 & 0.3223 & 0.2900 & 0.4826 & 0.3497 & 0.4614 & 0.3965 & 0.3384 & 0.2273 & 0.0591 & 0.1829 \\
SI-EN & 0.2902 & 0.2298 & 0.2876 & 0.2245 & 0.3705 & 0.2737 & 0.2831 & 0.2162 & 0.3879 & 0.3104 & 0.3740 & 0.2862 & 0.2883 & 0.2086 & 0.0476 & 0.2169 \\

\hline
\multicolumn{17}{c}{ \textbf{Llama-2-13B}   }\\
 \hline

EN-DE & 0.0612 & 0.0436 & 0.0111 & 0.0082 & 0.0527 & 0.0229 & 0.0467 & 0.0534 & 0.0246 & 0.0874 & 0.0386 & 0.0735 & 0.0337 & 0.0266 & 0.0018 & 0.0487 \\
EN-MR & -0.0025 & -0.0021 & -0.0324 & -0.0248 & 0.026 & 0.0501 & -0.0062 & 0.0143 & 0.0341 & 0.0824 & 0.0506 & 0.0757 & 0.0957 & 0.0607 & 0.0349 & 0.0120 \\
EN-ZH & 0.0085 & 0.0067 & 0.0086 & 0.0067 & 0.0889 & 0.1033 & -0.0201 & 0.0593 & 0.0496 & 0.0744 & 0.0486 & 0.0699 & 0.0455 & 0.0521 & 0.0657 & 0.0628 \\
ET-EN & 0.2339 & 0.1821 & 0.1992 & 0.1440 & 0.3263 & 0.3051 & 0.0239 & 0.1603 & 0.0542 & 0.2841 & 0.088 & 0.3112 & 0.2104 & 0.1875 & -0.031 & 0.1707 \\
NE-EN & 0.0619 & 0.0492 & 0.3039 & 0.2229 & 0.2466 & 0.2359 & 0.0538 & 0.0842 & 0.0865 & 0.0933 & 0.038 & 0.2045 & 0.3043 & 0.2448 & 0.0385 & 0.1406 \\
RO-EN & 0.3048 & 0.2332 & 0.2916 & 0.2141 & 0.4189 & 0.3422 & 0.0302 & 0.1975 & 0.0698 & 0.2414 & 0.0463 & 0.3666 & 0.3420 & 0.2538 & -0.0267 & 0.1936 \\
RU-EN & 0.404 & 0.3118 & 0.1973 & 0.1442 & 0.3916 & 0.3021 & 0.0843 & 0.2463 & 0.0562 & 0.2331 & 0.3697 & 0.3148 & 0.2766 & 0.1899 & 0.0645 & 0.1493 \\
SI-EN & 0.0295 & 0.0212 & 0.3414 & 0.2501 & 0.2160 & 0.1992 & 0.0726 & 0.0342 & -0.063 & 0.0731 & 0.0705 & 0.2622 & 0.2478 & 0.2116 & 0.0827 & 0.1034 \\

\hline
\multicolumn{17}{c}{\textbf{Qwen1.5-14B}    }\\
 \hline

 EN-DE & 0.1555 & 0.1219 & 0.2543 & 0.1975 & 0.1504 & 0.1625 & 0.0717 & 0.1173 & 0.0143 & 0.1652 & 0.2114 & 0.1746 & 0.2089 & 0.1777 & 0.0254 & 0.1365 \\
EN-MR & 0.2572 & 0.2053 & 0.3300 & 0.2634 & 0.4550 & 0.2395 & 0.0407 & 0.1551 & 0.0161 & 0.2120 & 0.4507 & 0.2332 & 0.4174 & 0.2648 & 0.1059 & 0.2114 \\
EN-ZH & 0.3839 & 0.2914 & 0.2615 & 0.1976 & 0.3655 & 0.3028 & 0.0557 & 0.2245 & 0.0297 & 0.2497 & 0.4011 & 0.3212 & 0.3580 & 0.2479 & 0.0038 & 0.2154 \\
ET-EN & 0.5134 & 0.3867 & 0.5896 & 0.4510 & 0.5923 & 0.4796 & 0.0769 & 0.3477 & -0.0069 & 0.4063 & 0.1954 & 0.4871 & 0.5215 & 0.4336 & 0.0487 & 0.3170 \\
NE-EN & 0.3265 & 0.2376 & 0.5208 & 0.3857 & 0.5149 & 0.3741 & 0.1401 & 0.2391 & 0.1960 & 0.3282 & 0.4868 & 0.3338 & 0.4712 & 0.3578 & 0.0484 & 0.2395 \\
RO-EN & 0.5609 & 0.4387 & 0.5621 & 0.4348 & 0.7127 & 0.5527 & 0.0985 & 0.4176 & 0.0466 & 0.4385 & 0.7293 & 0.5345 & 0.7369 & 0.5479 & 0.1085 & 0.3300 \\
RU-EN & 0.5047 & 0.3982 & 0.4997 & 0.3979 & 0.6002 & 0.4332 & 0.1075 & 0.3620 & 0.1665 & 0.4161 & 0.6100 & 0.4282 & 0.5546 & 0.3957 & 0.3920 & 0.3728 \\
SI-EN & 0.3693 & 0.2669 & 0.5096 & 0.3773 & 0.5820 & 0.4366 & 0.0609 & 0.2002 & 0.1472 & 0.3012 & 0.5249 & 0.3997 & 0.4340 & 0.3537 & 0.1646 & 0.3033 \\
\hline
\multicolumn{17}{c}{\textbf{Mixtral-8x7B-Instruct}}\\
 \hline

EN-DE & 0.1980 & 0.1444 & 0.1658 & 0.1195 & 0.0189 & 0.1959 & -0.0161 & 0.1137 & -0.0713 & 0.1388 & 0.0404 & 0.1709 & -0.0827 & 0.1625 & 0.2556 & 0.1631 \\
EN-MR & 0.1819 & 0.1405 & 0.2115 & 0.1539 & 0.0186 & 0.1394 & 0.0140 & 0.0634 & 0.0230 & 0.1493 & 0.0707 & 0.1469 & 0.0025 & 0.1428 & 0.0299 & 0.1917 \\
EN-ZH & 0.3324 & 0.2424 & 0.2927 & 0.2109 & 0.0890 & 0.2687 & -0.0062 & 0.1496 & 0.0654 & 0.1926 & -0.0061 & 0.2474 & 0.2443 & 0.1813 & 0.3079 & 0.2200 \\
ET-EN & 0.4748 & 0.3531 & 0.5552 & 0.4095 & 0.2111 & 0.4614 & 0.0380 & 0.2517 & 0.0738 & 0.3171 & 0.1534 & 0.4336 & 0.2656 & 0.3399 & 0.0567 & 0.4052 \\
NE-EN & 0.3909 & 0.2872 & 0.4215 & 0.3070 & 0.1190 & 0.3572 & 0.0543 & 0.0842 & -0.0219 & 0.1629 & 0.1416 & 0.3409 & 0.4209 & 0.3223 & 0.4284 & 0.3824 \\
RO-EN & 0.5576 & 0.4344 & 0.5416 & 0.4107 & 0.0427 & 0.5000 & 0.0748 & 0.3024 & 0.0991 & 0.3196 & 0.0947 & 0.4727 & 0.6549 & 0.4879 & 0.2054 & 0.5154 \\
RU-EN & 0.4823 & 0.3630 & 0.4514 & 0.3422 & 0.1760 & 0.4510 & 0.0854 & 0.3055 & 0.1199 & 0.3767 & 0.1844 & 0.4257 & 0.5816 & 0.3913 & 0.0212 & 0.3984 \\
SI-EN & 0.3352 & 0.2415 & 0.4459 & 0.3223 & 0.0961 & 0.3343 & 0.0072 & 0.0691 & 0.0934 & 0.1555 & 0.1099 & 0.2997 & 0.4255 & 0.3461 & 0.4334 & 0.3441 \\ 
\bottomrule
\end{tabular}%
}
\caption{Pearson's $r$ and Kendall's $\tau$ correlation scores achieved using Templates 1-8 (T1-8) on various
open-source LLMs for each language pair (LP).}
\label{tab:results_openchat}
\end{table*}

\begin{CJK*}{UTF8}{gbsn}
\begin{figure*}[ht]
\centering
  \begin{minipage}{12cm}
    \textbf{Model input} (before formatting): \\
    Score the following translation from English to Chinese with respect to the human reference on a continuous scale from 0 to 100, where score of zero means "no meaning preserved" and score of one hundred means "perfect meaning and grammar".\textbackslash nEnglish source: The last conquistador then rides on with his sword drawn.\textbackslash nChinese human reference: 最后的征服者随后举着剑前进。\textbackslash nChinese translation: 最后的征服者骑着他的剑继续前进.\textbackslash nScore: \\
    \textbf{Model output:} \\
    <|im\_start|>user\textbackslash nScore the following translation from English to Chinese with respect to the human reference on a continuous scale from 0 to 100, where score of zero means "no meaning preserved" and score of one hundred means "perfect meaning and grammar". \textbackslash nEnglish source: The last conquistador then rides on with his sword drawn.\textbackslash nChinese human reference: 最后的征服者随后举着剑前进。\textbackslash nChinese translation: 最后的征服者骑着他的剑继续前进.\textbackslash nScore:<|im\_end|>\textbackslash n<|im\_start|>assistant\textbackslash n最后的征服者骑着他的剑继续前进.
  \end{minipage}
  \caption{An example of the TowerLLM output for scoring English-to-Chinese translation using Template 3 via HuggingFace. The output was generated in March, 2024.}
\label{towerllm}
\end{figure*}
\end{CJK*}

\clearpage
\end{document}